\PassOptionsToPackage{hypertexnames=false}{hyperref}

\documentclass{sn-jnl}

\usepackage{graphicx}%
\usepackage{multirow}%
\usepackage{amsmath,amssymb,amsfonts}%
\usepackage{amsthm}%
\usepackage{mathrsfs}%
\usepackage[title]{appendix}%
\usepackage{xcolor}%
\usepackage{textcomp}%
\usepackage{manyfoot}%
\usepackage{booktabs}%
\usepackage{algorithm}%
\usepackage{algorithmicx}%
\usepackage{algpseudocode}%
\usepackage{listings}%
\usepackage{tipa}
\usepackage[authoryear]{natbib}
\usepackage{geometry}
\usepackage{tikz}
\usetikzlibrary{arrows.meta, positioning, shapes.geometric, shadows}

\begin{document}

\title[Article Title]{Figurative Justice: Detecting metaphors in Hindi judgements with qualitative assessment and transformers

}


 \author*[1]{\fnm{Bhumika} \sur{Bhattacharyya}}\email{bhumika.bhattacharyya@ut.ee}

 \author[2]{\fnm{Shouvik Kumar} \sur{Guha}}\email{shouvikkumarguha@nujs.edu}

 \author[3]{\fnm{Indranil} \sur{Dutta}}\email{indranildutta.lnl@jadavpuruniversity.in}

 \affil*[1]{\orgdiv{DigiTS}, \orgname{University of Tartu}, \country{Estonia}}

 \affil[2]{\orgdiv{Department}, \orgname{The West Bengal National university of Juridical Sciences (NUJS)}, \country{India}}

 \affil[3]{\orgdiv{School of Languages and Linguistics}, \orgname{Jadavpur University}, \country{India}}


\abstract{Metaphors are figurative use of words for conceptual mapping \citep{lakoff1981}. Metaphor detection in the legal context has been crucial as metaphors are  persuasive juridical means of creating legal meaning and concepts resulting in significant consequences. Metaphorical framing in legal discourse by judges, lawyers, and legislators brings about real-time implications upon individuals (\citep{Bozovic2024}) and influences judicial decision-making, argumentation and interpretation of laws. This is crucial in Human Rights infringement cases where language determines severity of punishment, public perception and judicial outcomes.

While automatic metaphor detection in major languages like English, Spanish, Polish, Lithuanian have aided in understanding inherent intentions of metaphorical use of language \citep{sanchez-boyona2025, WWP_2019,Urbonaite2017}, there is no such attempt in low-resource languages like Hindi. The dearth of annotated legal corpora in Hindi makes it difficult to develop NLP models and detect metaphors in judicial proceedings. In the Indian context, Convolutional Neural Networks (CNNs) have been used for classification of bail judgements \citep{barman2023}, however there are no existing models designed for metaphor detection.

We present a Hindi Legal Metaphor Corpus (HiLeMe) by isolating judgements from Hindi Legal Data Corpus (HLDC) \citep{kapoor-etal-2022-hldc}. Legal experts annotated HiLeMe to classify metaphorical constructions using the MIPVU schema \cite{steen2010}. We downstreamed an mBERT on Hindi legal metaphor detection task \cite{devlin2019}. We built a transformer-based architecture for metaphor detection that are known to outperform traditional models in legal classification tasks \citep{imran2023} This model provides insights into the judicial psyche for decoding judicial decisions. Our research contributes to advancing automated models in legal discourse in low-resource languages like Hindi and envisages adoption into 22 Indian schedule languages.
}

\keywords{Legal Metaphors,AI,Hindi}



\maketitle

\section{Introduction}\label{sec1}

\subsection{Metaphors and figurative usage}
In their seminal work, \cite{lakoff1981} provide a comprehensive account of what makes metaphorical usage in natural language interesting in order to put forward a theory of conceptual metaphors. Perhaps motivated by generative linguistics, \cite{lakoff1981} argue for conceptual metaphors as underlying forms for surface metaphorical usage. Interpretation of metaphors, therefore, relies heavily on unearthing their conceptual and underlying forms.  \cite{steen2010} provide a critique of conceptual metaphors on several accounts; lack of constraints in delimiting what count as conceptual metaphors, difficulty in identification of linguistic terms that may be related to conceptual metaphors, and general evolution of the idea of conceptual metaphors. \cite{steen2010} put forth a framework, Metaphor Identification Procedure Vrije Universiteit (MIPVU) which is an iteration of the earlier MIP \citep{PragglejazGroup2007}. 

While identification of metaphors is a non-trivial task in domain general texts, identifying metaphorical usage in legal discourse is further complicated. Use of metaphorical and figurative language in law and legal discourse has both salubrious and deleterious effects \citep{Ebbesson2008}. Use of metaphors in legal language aid and facilitate understanding somewhat obtuse legal concepts by way of analogy. At the same time, metaphorical usage can lead to adoption of ideas uncritically and perhaps to an extent outside the bounds of what may be possible within the limits of interpretability of certain legal positions and concepts. \cite{Ebbesson2008} also points to a crucial characteristic of metaphors - the ability to polarize opinion; one may develop positive or negative associations with metaphors. In that sense, metaphorical usage in legal discourse edges closer to rhetorical usage, acceptance of the same may lead to changed outcomes in legal matters. 

\cite{Urbonaite2017} is an extensive attempt to look at metaphorical usage in legal discourse from a cross-linguistic perspective. It is important to stress here and preface computational approaches with the inherent difficulties of metaphorical identification. Unlike other models of language, large of otherwise, metaphorical understanding and reasoning is not always compositional in terms of meaning and therefore interpretation of metaphors in one language doesn't necessarily aid interpretation in another. Essentially, artificial learning of metaphorical usage is not language agnostic and lends itself with some difficulty to transfer learning \citep{bozinovski2020}. This can be seen even in simple examples; use of ``defendant'' signifies the underlying process of court proceedings akin to ``war'' in English. However, in Hindi, for example, the use of the term [\textipa{{\textprimstress pr@t\textsubbridge Ivad\textsubbridge I}}
] for ``defendant" has the implication of someone holding an opposing view and doesn't therefore necessarily invoke the ``argument is war" notion, at least not in the most obvious of senses. 

In section \ref{sec:qa}, we discuss in greater detail various issues that surround a qualitative understanding of metaphorical use in legal discourse. Following that, in section \ref{sec:autodetect}, we review the various attempts at automatic detection of metaphors. In section \ref{sec:model}, we outline our approach and discuss in detail the transformer model we used to train and test on annotated metaphors on the HLDC corpus \citep{kapoor-etal-2022-hldc}. We also discuss the various components of the Hindi Legal Metaphor (HiLeMe) corpus, especially the use of MIPVU in labeling HiLeMe. In section \ref{sec:arch}, we outline our model architecture , train a BERT based multilingual cased model for metaphor detection task and evaluate it's performance. In section \ref{results}, we present our overall model performance and discuss the implications of our results for further building language agnostic metaphor detection models for schedule languages in India.

\section{Qualitative assessments of metaphorical usage in legal discourse}\label{sec:qa}
The positivist image of law as a domain of exacting precision, a science of rules and syllogisms where language serves as a neutral vessel for the transmission of legislative will and judicial logic, may not leave any room for linguistic ambiguity, yet it is undeniable that the law is inescapably, structurally, and pervasively metaphorical. Metaphors do not play any mere decorative role or rhetorical flourish in legal language, rather they are an integral part of the lens through which lawyers, judges and legal scholars seek to conceptualize abstract normative principles and translate them into operational realities. Precedents, doctrines and constitutional principles are at times framed by vivid images of identities, penumbras, trees, walls, markets, weapons, and many others. These metaphors simultaneously enable legal reasoning and conceal contestable normative choices. An examination of the trend of usage of metaphors in the legal domain is likely to provide a core insight, viz. law cannot do without metaphors, but unexamined metaphors can mislead, entrench power and distort adjudication. Instead of seeking to purge or purify law of metaphors, the end goal ought to be to make their operation visible, accountable and open to critical review, be it in any type of legal document.
The necessity of metaphor arises from the very abstract nature of legal concepts like rights, duties, jurisdiction, justice, negligence, and sovereignty, which have no physical existence and hence to manipulate these abstractions, the human mind must map them onto concrete, sensory domains of experience. Hence the legal attempts to understand ``jurisdiction" through the metaphor of physical territory (``long-arm" statutes that ``reach"), to understand ``argument" through the metaphor of war (positions are ``attacked" and claims ``defended" claims), or to understand ``justice" through the metaphor of physical balance (the ``scales of justice"). As the philosopher Giambattista Vico argued, and as legal scholars like G. Joseph Vining and James Boyd White have echoed, this process is not merely linguistic but ontological - we create the legal world by naming it through metaphor \footnote{Edward L. Murray, The Phenomenon of the Metaphor: Some Theoretical Considerations, in A. Giorgi, C. Fischer \& E. Murray (eds.), Duquesne Studies in Phenomenological Psychology, vol. II, 288 (Pittsburgh, 1975).}. 

Yet this symbiotic relationship between law and metaphor is fraught with tension. As Benjamin Cardozo had warned, metaphors in law are ``to be narrowly watched, for starting as devices to liberate thought, they end often by enslaving it".\footnote{Michael Frost, Greco-Roman Analysis of Metaphoric Reasoning, 2 Legal Writing: The Journal of the Legal Writing Institute 113 (1996).} This is likely to occur when the metaphorical nature of a legal concept is forgotten, and the metaphor is instead mistaken for the reality it represents. For instance, when a court treats a company not just as if it were a person for the purpose of a contract, but as a person with right to privacy, the metaphor gets hardened into a doctrine that may obscures the original policy rationale.

James Murray, while examining the hypothesis that judicial opinions``work the way a metaphor works," posited law not as a set of objectively derivable doctrinal formulae, but as a process that generates precedents operating as persuasive moral injunctions. \footnote{James E. Murray, Understanding Law as Metaphor, 34 Journal of Legal Education 714 (1984).} Drawing on interaction and tension theories of metaphor, as propounded by the likes of Max Black, Heidegger and Vico, Murray argued that legal concepts like ripeness, standing and mootness are not natural but metaphorical constructs. \footnote{\emph{Id.}} Murray also opined that metaphor is cognitive, hence capable of creating new legal meanings. When the judge is ``naming" a new right or doctrine, they are engaging in metaphorical world‑building. Murray’s position therefore aligns with the cognitive-linguistic framework popularised by \cite{lakoff1981}, which treats conceptual metaphor as a fundamental way of structuring abstract domains (for instance, ‘life is a journey’). If one applies similar principles in the legal domain, it may lead to models like ‘law is a structure’, or ‘rights are objects’ – said models can then shape the way legal players reason, often beneath conscious awareness.

In persuasive legal writing, metaphors have been found to occur at multiple levels, as discussed by Michael Smith \footnote{Michael R. Smith, Levels of Metaphor in Persuasive Legal Writing, 58(3) Mercer Law Review 919 (2007).}, viz. Doctrinal Metaphors (metaphor as binding law, such as ``marketplace of ideas \footnote{This transforms free speech jurisprudence into a quasi-economic analysis where ``competition" determines truth and state intervention is viewed as ``market distortion."}," ``wall of separation," ``fruit of the poisonous tree \footnote{This vivid botanical image governs the admissibility of evidence: if the primary evidence (the tree) was obtained illegally, any secondary evidence derived from it (the fruit) is equally tainted and must be excluded.}," ``piercing the corporate veil \footnote{The ``veil" is a fictitious barrier separating the legal personality of the company from its shareholders. Courts do not ask whether the statutory requirements for liability are met in abstract terms. They ask whether the ``veil" should be ``pierced", ``lifted", or ``torn". This textile imagery structures the entire inquiry into limited liability.}," ``standing," ``long‑arm jurisdiction"), Legal Method Metaphors (metaphor explaining the mechanics of adjudication through images used to describe how courts reason, such as ``balancing" interests \footnote{This metaphor draws on the iconography of Lady Justice holding scales. It suggests a precise process where competing abstract values (such as privacy v. security) are placed on pans to determine which is heavier. However, unlike physical objects, legal interests have no objective mass or weight. When a judge states that public order “outweighs” freedom of assembly, they are not reporting a measurement. Rather, they are making a subjective, normative value judgment. The metaphor therefore serves a legitimizing function, cloaking discretionary political choices in the language of objective measurement and scientific precision. It implies that the judge is merely reading a scale rather than making a choice.}, ``opening the floodgates," ``finding" law in ``sources"\footnote{The positivist tradition often portrays the judge as an archaeologist or explorer who ``finds" the law within the ``sources" (itself a hydraulic metaphor of springs and rivers). This suggests that the legal rule exists pre-formed, waiting to be discovered. In reality, judges often ``manufacture" fresh legal rules (or at least fresh interpretation of rules) to address novel situations. The ``finding" metaphor obscures this creative agency, protecting the judiciary from charges of activism by framing their work as passive discovery.}), Stylistic Metaphors (conscious rhetorical devices using vivid analogies in opinions and advocacy, such as copyright infringement being called ``piracy"\footnote{This is a stylistic choice designed to stigmatize the defendant by associating them with violent maritime robbery.}, regulations having ``chilling effect"\footnote{This invokes a thermal metaphor to suggest an imperceptible but dangerous dampening of rights.}), and Inherent Metaphors (normalised legal vocabulary having oft-forgotten metaphorical nature, such as ``duty," ``burden," ``rights" as things that can be ``carried" or ``waived", ``breaking" a contract, a ``binding" agreement, or a ``cause of action"\footnote{These metaphors rely on physical concepts of force, connection, and causality to structure basic legal relationships.}).

A holistic view of the above may lead one to conclude that when studying or practising law, it is ``metaphor all the way down", leaving one only to decide not between metaphor and literalism but between better and worse metaphors.

Previous studies of metaphorical use in legal judgements emanating from the European Court of Justice reveal that metaphors are not incidental flourishes, but rather systematic conceptual patterns. Following \cite{Bozovic2024}, one may identify recurrent conceptualisations of key legal domains such as law being a person (law ``requires," ``governs," ``confers", it has a body as reflected by ``corpus juris", with functional limbs as shown by ``the long arm of the law", and even reproductive capacity with seminal cases having their ``progeny"). Similar instances may be cited shaping concepts like law as a sacred place (rights and principles ``enshrined" in law), law being a source of light (``examined in the light of the law"), law being like a structure or tree (with ``branches," ``foundations" and ``frameworks," suggesting both stability and growth) and so on. Courts may be subject to similar personification (they ``hold," ``find," ``ask", they are spatially ordered as higher/lower courts, reinforcing vertical authority). According to \cite{Bozovic2024}, these metaphors form ``conceptual clusters," not isolated images. \footnote{\emph{Id}}

On the other hand, \cite{kyomugisha2025} offers a more recent survey of ``legal metaphors and justice" and combines cognitive linguistics with critical discourse analysis, distinguishing between system‑building conceptual metaphors (law as architecture, system, structure), which presuppose coherence and rational derivation; and linguistic metaphors about actors and events (``the law is an ass"), which signal attitudes towards institutions and outcomes. \footnote{ Jonas Ebbesson, Law, Power and Language: Beware of Metaphors, 53 Scandinavian Studies in Law 259 (2008).}

A closer analysis of similar metaphorical usage in judgements is likely to reveal the judiciary and even other stakeholders in the judicial process may share an inclination towards metaphors albeit to varying degrees. If that is indeed true, then changing the dominant metaphor in a line of cases may correlate with shifts in doctrinal outcomes. This underpins the call for an ``ethical and deliberate" use of metaphor, rather than the naïve assumption that metaphor is merely stylistic decoration.

Studies like \cite{Ebbesson2008} go far in establishing the link between figurative language and the power of lawyers. Figurative expressions are everywhere in legal practice, not merely in vivid labels (``defence," ``higher" courts, ``war"‑like adversarial procedure) but in core concepts such as ``sources" of law, ``balancing" of interests and ``legal system." \cite{Ebbesson2008} believes that the metaphor of ``finding" answers in ``sources" of law depicts judges as passive discoverers, masking their constructive role in developing the law. Yet he does not call for abandoning metaphors, rather he insists that ``metaphor is the traditional device of persuasion."\footnote{\emph{Id}} The normative demand that he makes, following in the footsteps of Fuller, is instead to ``see and see through" metaphors, especially when they legitimate the authority and exclusivity of legal professionals.\footnote{\emph{Id}} This approach gets further reinforced by studies reflecting how metaphors encode assumptions about social hierarchy, gender and race, for instance in rape‑as‑war metaphors or in metaphors that naturalise victims’ suffering.\footnote{ Asiimwe Kyomugisha T., Analyzing Legal Metaphors: Implications for Justice, 4(2) Research Invention Journal of Current Issues in Arts and Management 1 (2025).} Such metaphors can subtly tilt adjudication against vulnerable groups, even when doctrinal standards appear neutral. 

Stephen Newman’s work on stylistic metaphors in the American appellate practice offers enriching insights in this regard, with decisions that must be ``as wrong as a five‑week‑old unrefrigerated dead fish," private homes turned into ``soundstages" for police theatricals, or license plates used as ``mobile billboards" for state ideologym\cite{Newman1999}. \footnote{Stephen A. Newman, Uses of Metaphor in Legal Argument, 222 New York Law Journal 1 (1999).} These images, while capable of communicating complex legal evaluations quickly, also pose the danger of tacitly raising or lowering proof thresholds without doctrinal change.\footnote{An expert’s comparison of memory to a computer ``button" in a rape trial overstates the reliability of human recall, and a ``jigsaw puzzle" metaphor for circumstantial evidence implies a standard of seamless fit that no real case can meet. } \cite{smith2007} carries this discussion further, by treating doctrinal metaphors as targets of ``Cardozo attack", demonstrating that a metaphor (``wall of separation" between church and state) no longer captures contemporary realities, and uses that critique to justify more nuanced tests (such as the \textit{Lemon} three‑part test).\footnote{Michael R. Smith, Levels of Metaphor in Persuasive Legal Writing, 58(3) Mercer Law Review 919 (2007).}

The relationship between metaphors and law gains greater substance from existing doctrinal case studies such as those by \cite{Lewison2015}, \cite{MohseniGummow2024} of metaphors that have caused real legal error. \footnote{A few examples include literal acceptance of metaphors (forfeiture clauses as “security for rent” read literally into insolvency statutes, causing judicial treatment of landlords as “secured creditors” in a sense unintended by the legislature), metaphors used as conclusory labels (phrases like “breach going to the root of the contract” functioning as botanical metaphors but obscuring significant underlying questions about whether the breach deprives the innocent party of substantially the whole or part of the contractual benefit), and anthropomorphic metaphors masking policy (the “directing mind and will” of a company, or the comparison of corporations to human bodies with brains and hands, has caused judicial treatment of complex attribution questions like a quest seeking an individual puppet‑master).} In particular, in constitutional law studies, metaphors often tend to play an even greater role by expressing visions of the polity, mediating between domestic and supranational orders, and framing identity conflicts. This is a phenomenon observed across jurisdictions, such as the Canadian ``living tree" Constitution and the Australian ``skeletal principle" metaphors illustrating the risk that constitutional imagery may become a ritual incantation rather than an aid to reasoning, with the judiciary having to insist that while the \textit{denotation} of constitutional terms can extend to new phenomena (such as same‑sex marriage), their \textit{connotation} cannot be re‑written by metaphor alone \cite{MohseniGummow2024}. When it comes to the Indian Constitution, there are multiple examples of metaphorical usage shaping core constitutional principles - the Constitution has been described as a ``living document," ``the vehicle of the life of a nation," ``not a gate but a road" (grounding a dynamic, purposive interpretive style), with penumbral rights emanating from express guarantees (like the recognition of privacy as a penumbral facet of Article 21), the treatment of the ``basic structure" as a quasi‑sacred core (a structural metaphor having acquired an almost theological aura of inviolability), doctrines such as ``constitutional morality" being considered as a standard with ``a value of permanence," the phrase ``We, the People" being considered as a metaphor for popular sovereignty whose scope has expanded over time, and the subsequent insertion of ``socialist" into the Preamble, now judicially read as a metaphor for equality of status and opportunity rather than as a rigid endorsement of state ownership. \footnote{Shivangi Gangwar and Aishwarya Pagedar, Examining the Living Metaphor in the Indian Constitution, 13(2) Jindal Global Law Review 347 (2022). \textit{See} also Apurva Mittal and Vishavjeet Chaudhary, Law and Literature: Interpretation of the Constitution, 1 Bennett Journal of Legal Studies 151 (2020). See also Saai Sudharsan Sathiyamoorthy, Morality, Metaphors and the Constitution, August 23, 2024, The New Indian Express, available at https://www.newindianexpress.com/opinions/2024/Aug/22/morality-metaphors-and-the-constitution (last visited November 26, 2025).} While these constitutional metaphors have enabled a remarkable expansion of rights without formal textual amendment, they also carry an element of risk. \footnote{With seemingly contrasting metaphors such as “living document” and permanence both at its disposal, the courts can oscillate between progressive evolution and conservative immutability, invoking one metaphor or the other to justify contested outcomes. Similarly, treating basic structure as a sacred “core” can discourage granular analysis of particular amendments and foster a binary mentality (constitutional v. unconstitutional) that reduces political space for legislative experimentation. If “constitutional morality” is not anchored in publicly reasoned principles but in judicial sensibilities, its metaphorical elevation can mask moral disagreement and fuel accusations of judicial overreach. See Shivangi Gangwar and Aishwarya Pagedar, Examining the Living Metaphor in the Indian Constitution, 13(2) Jindal Global Law Review 347 (2022).} 

Therefore, based on the aforesaid discussion, one may be tempted to draw an inference that responsible use of legal metaphors may require certain guardrails to be established, such as making metaphors visible and contestable (judicial clarification about their limited function rather than doctrinal treatment), avoiding metaphors being perceived as self-standing rules, auditing commonly used metaphors for distributive impact, and impart training, education, and awareness about metaphors and their significance to present and future legal professionals, to name a few.


\section{Complexities of automatic detection of Metaphors}
\label{sec:autodetect}
Automatic detection of metaphors both in domain general and in the legal domain has had somewhat limited success, unsurprisingly due to the aforementioned complexities, nuances, and also language specific constraints in understanding and interpreting all types of metaphors. I the is section, first we will provide a brief summary of domain general computational approaches to metaphor identification and detection. Following that we will look into legal domain specific computational methods, specifically human rights cases but not necessarily only in that legal domain.

\subsection{Metaphor detection in NLP applications}
Within broader domain general approaches both unsupervised and supervised methods for metaphor detection can be found. \cite{pramanick-mitra-2018} proposes an unsupervised framework that trains on adjective-noun (AN) pairs from a dataset of that consists of both literal and metaphorical adjective-noun pairs \citep{tsvetkov2014}. The unsupervised k-means clustering achieved an accuracy of 72\% in \cite{pramanick-mitra-2018}, however, their results couldn't be compared with \cite{tsvetkov2014} because of the differences in methods; \cite{tsvetkov2014} reported an F-score of 0.85. Interestingly, the model trained by \cite{tsvetkov2014} on English was also simultaneously tested on three other languages; Farsi, Russian and Spanish with varying degrees of success on the AN task - the reported f-scores where between 0.72 for Spanish, 0.74 for Farsi and 0.77 for Russian.

More recent approaches have used pre-trained language models for metaphor detection and identification \citep{aghazadeh-etal-2022}. Approaches based on pre-trained models offer the added advantage of determining the representational architecture of metaphors in the pretrained layers of the models. In their approach, \cite{aghazadeh-etal-2022} probe the representational structures in the models and claim that metaphorical information is encoded as contextual representations. In the last few years, the State-of-the Art models, therefore, have exclusively used pre-trained models such as BERT and their iterations \citep{devlin2019}. 

So far though, all of these approaches were mostly applied on languages such as English, Russian, Spanish and Farsi. Low resource languages have continued to pose challenges for these tasks due to lack of pre-trained models, datasets and adequate information-rich labeling. \cite{schneider-etal-2022-metaphor} present an unsupervised approach, again on adjective-noun pairs and address the lack of resources by detecting metaphors from AN pairs in Middle High German. The methods employed in \cite{schneider-etal-2022-metaphor} include measuring cosine distances between literal and metaphorical AN pairs, where the latter exhibit greater distances in the transformed vectors. \citep{jia2024} propose a prompt learning method by which they guide the model to come up with contextual meanings so as to mitigate the influence literal interpretations have. All things considered, we come to understand that despite the success of BERT based architectures in metaphor identification and representation - research on under-studied or lesser known languages is sparse.

\subsection{Metaphor detection in the legal domain}

Within the legal domain, model based identification of metaphors has not gotten much traction. There are of course several studies that have employed transformer based models, albeit for judgment classification; \cite{barman2023,kapoor-etal-2022-hldc,imran2023}. The state-of-the-art techniques have therefore been limited to classification of judicial documents \cite{imran2023}, prediction of bail judgments \cite{barman2023} and if developed on oft studied languages have built summarization systems. Automatic detection of metaphorical usage in under-studied languages remains a completely unchartered territory.

In what follows, we outline our methods to use MIPVU \citep{steen2010} to annotate the Hindi Legal Data Corpus (HLDC; \citep{kapoor-etal-2022-hldc}) to create a Hindi Legal Metaphor (HiLeMe) corpus. We further use BERT Base Multilingual Cased model which was fine-tuned on a metaphor detection task.

\section{Model pipeline and our approach}
\label{sec:model}
Given the lack of computational resources for Hindi figurative language, we have developed a pipeline for a supervised model to detect metaphoric use of lexical units in Hindi legal discourse. 

\subsection{Data}
Hindi sentences were extracted from the HLDC \citep{kapoor-etal-2022-hldc} corpus that contains district court judgments from 71 districts of the state of Uttar Pradesh in India. These judgments are exclusively written in Hindi and therefore act as both a challenge and an opportunity to test how language agnostic models may in fact work in this particular use case. The districts were chosen on the basis of a higher number of cases in which most of the data came from Saharanpur, Sitapur, Ghaziabad, Muzaffarnagar and Agra (all district level courts in the Indian state of Uttar Pradesh). These cases were then programatically split into sentences for identification of lexical units in the MIPVU annotation \citep{steen2010}. The dataset supporting this study is available in the Open Science Framework repository: https://osf.io/z398e/

\subsection{Data Availability}

The dataset supporting the findings of this study is available in the Open Science Framework repository at \url{https://osf.io/z398e/overview?view_only=76ec6a5d7da444cf8bfb6e58c79b1ccd}.

\subsection{Annotations}
Linguists trained six legal experts for MIPVU annotations. Each annotator received a total of 200 cases divided into sentences. Annotators were then asked to independently apply domain knowledge on each lexical unit, identifying contextual meaning and basic meaning of the chosen lexical unit, providing a decision as to Yes/No (metaphor/no metaphor), a rationale for their decision, and confidence.

\subsection{Preprocessing}
All annotated files were concatenated into one single dataframe. A few steps of data cleaning and standardization was done on the dataframe which included sentence string normalisation, removal of noise markers and bracket stripping. Missing or blank lexical units were normalised as empty lists. On the basis of decision row as marked by the annotators, a new label row was created with 0/1. A sentence was labeled 1 (metaphor) if the annotator decision was ``Yes", and lexical units were present. Otherwise, label = 0 (non-metaphorical). Utilizing Stanza’s Hindi model, each sentence was tokenized and POS tags were obtained.

\subsection{HiLeMe}
Our final dataset contains a total of 7137 sentences, comprising 162196 tokens, with an average length of 22.73 tokens per sentence. Experts annotated 2022 sentences as containing a metaphorical lexical unit following the MIPVU criteria, and has been summarized in table \ref{tab:dataset_stats}. The dataset is skewed as No metaphors exceed Metaphors, which suggests the lower frequency of figurative use of language in legal discourse. From contribution of our six annotators who had identified lexical units, we have extracted lexical unit frequency.

\begin{table}[h!]
\centering

\begin{tabular}{l r}
\toprule
\textbf{Statistic} & \textbf{Value} \\
\midrule
Total sentences & 7137 \\
Total tokens & 162196 \\
Average tokens per sentence & 22.73 \\
Sentences with $\geq$ 1 metaphor & 2022 \\
\midrule
No metaphor (label = 0) & 5115 \\
Metaphor (label = 1) & 2022 \\
\bottomrule
\end{tabular}
\caption{HiLeMe Statistics}
\label{tab:dataset_stats}
\end{table}








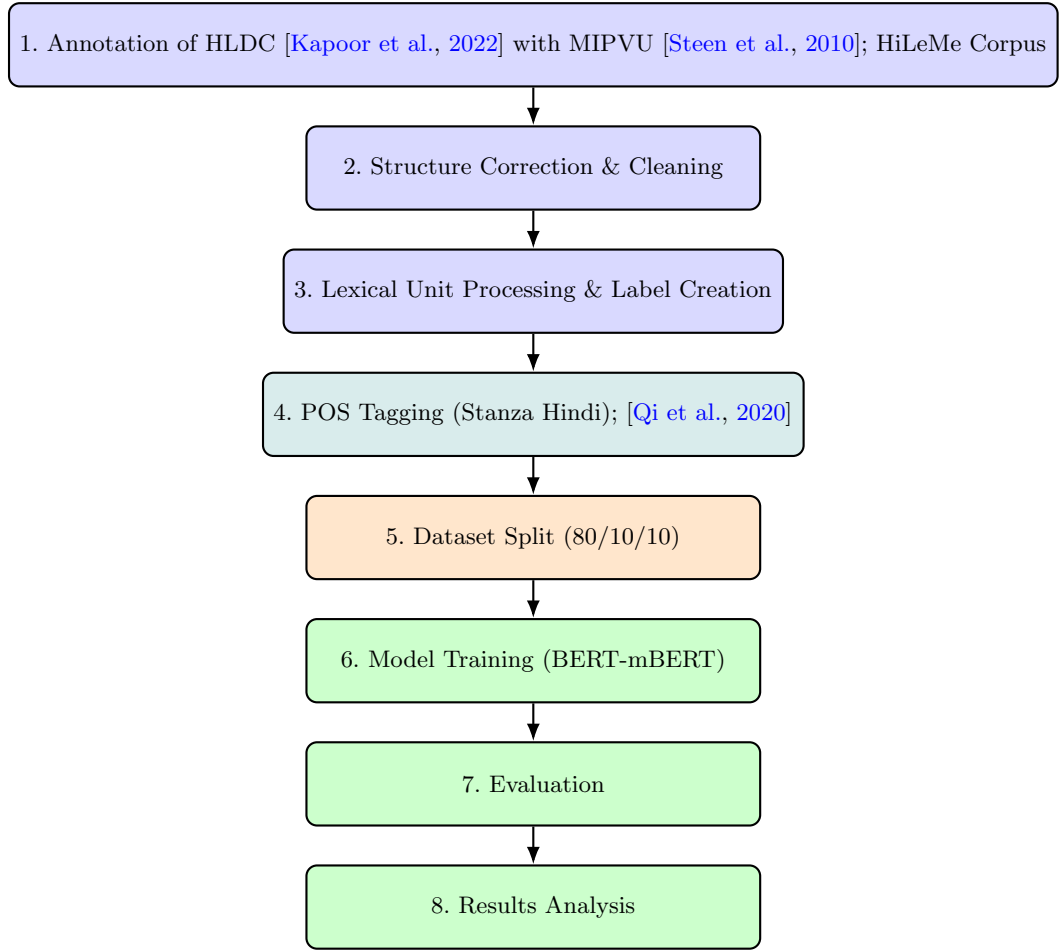
\begin{figure}[h!]
\centering
\begin{tikzpicture}[
    >=Latex,
    node distance=0.50cm,
    process/.style={
        rectangle,
        rounded corners,
        draw=black,
        thick,
        minimum width=6cm,
        minimum height=1.1cm,
        text centered,
        font=\small,
        inner sep=4pt
    },
    preproc/.style={process, fill=blue!15},
    ling/.style={process, fill=teal!15},
    split/.style={process, fill=orange!20},
    model/.style={process, fill=green!20},
    arrow/.style={->, thick}
]

\node[preproc] (n1) {1. Annotation of HLDC \citep{kapoor-etal-2022-hldc} with MIPVU \citep{steen2010}; HiLeMe Corpus};
\node[preproc, below=of n1] (n2) {2. Structure Correction \& Cleaning};
\node[preproc, below=of n2] (n3) {3. Lexical Unit Processing \& Label Creation};
\node[ling, below=of n3] (n4) {4. POS Tagging (Stanza Hindi); \citep{qi2020stanza}};

\node[split, below=of n4] (n5) {5. Dataset Split (80/10/10)};
\node[model, below=of n5] (n6) {6. Model Training (BERT-mBERT)};
\node[model, below=of n6] (n7) {7. Evaluation};
\node[model, below=of n7] (n8) {8. Results Analysis};

\draw[arrow] (n1) -- (n2);
\draw[arrow] (n2) -- (n3);
\draw[arrow] (n3) -- (n4);
\draw[arrow] (n4) -- (n5);
\draw[arrow] (n5) -- (n6);
\draw[arrow] (n6) -- (n7);
\draw[arrow] (n7) -- (n8);

\end{tikzpicture}

\caption{Pipeline for metaphor detection}
\label{fig:pipeline}
\end{figure}

\section{Model Architecture and Training}
\label{sec:arch}
\subsection{Model choice}

We have fine tuned a BERT-base-multilingual-cased model as represented in figure \ref{fig:pipeline} from Google  that was trained on 104 languages which includes Hindi. Our choice of model aligned with our goals as this model is primarily aimed at being fine-tuned on tasks that use the whole sentence to make decisions. Therefore, tasks like sentence-level figurative language use were close to its scope.

\subsection{Training}
For model training, we fine-tuned the BERT-base-multilingual-cased transformer using the sentence-level metaphor labels derived from the MIPVU annotations. The data set was stratified into 80-10-10 splits for training, validation, and testing to preserve the balance of the classes. The inputs were tokenized with a maximum sequence length of 128, and the model was optimized using AdamW with a learning rate of 2e-5, a batch size of 16, and three training epochs. Training was performed on a GPU-enabled Google Colab environment, with validation executed after each epoch to select the best-performing checkpoint. 

\subsection{Evaluation}

Model performance was assessed in the closed test set after the final training epoch. The fine-tuned mBERT model achieved an evaluation loss of 0.5280, with an inference speed of roughly 319.5 samples/second, indicating efficient runtime behavior. Overall accuracy reached 71.99\%, reflecting the model’s ability to distinguish literal from metaphorical usage in judicial text. For the metaphor class specifically, the model obtained a precision of 0.5077, recall of 0.3267, and an F1-score of 0.3976, suggesting that while the classifier is reliable in identifying clear metaphorical cues, it remains conservative in detecting more subtle metaphorical expressions. Our results, summarized in table \ref{model performance table},  align with common challenges in metaphor detection, specifically in domain-specific, low-resource legal contexts where figurative language is generally nuanced and highly dependent on the context.

\begin{center}
    
\begin{table}[h!]

\resizebox{0.3\textwidth}{!}{
\begin{tabular}{l c}
\hline
\textbf{Metric} & \textbf{Score} \\
\hline
Accuracy & 0.7199 \\
\hline
Precision & 0.5077 \\
Recall  & 0.3267 \\
F1-score  & 0.3976 \\
\hline
\end{tabular}}
\caption{Model performance on overall accuracy and metaphor class metrics}
\label{model performance table}
\end{table}
\end{center}

\begin{center}
    \begin{figure}[h!]
        \centering
        \includegraphics[width=0.5\linewidth]{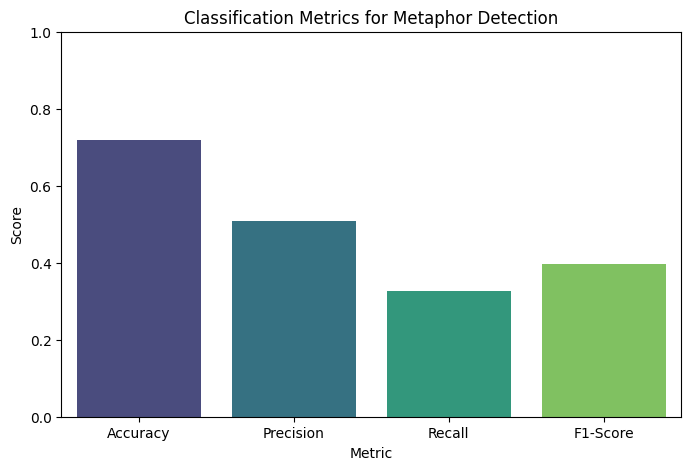}
        \caption{Metrics showing the model’s performance on accuracy, precision, recall, and F1-score.}
        \label{fig:model performance bar graph}
    \end{figure}
\end{center}

As can be seen in figure \ref{fig:model performance bar graph}, while the overall accuracy is at 71.99\% the precision, recall, and F1-scores are close to 0.5, 0.33, and 0.39, respectively. These results suggest a closer inspection of the false positives and negatives.

\section{Results}\label{results}
This section summarizes the performance of a fine-tuned mBERT model on a Hindi legal metaphor detection task. The model achieved an overall accuracy of 0.7199. For the 'metaphor' class, the model demonstrated a precision of 0.5077, recall of 0.3267, and an F1-score of 0.3976. These metrics suggest that there still is a great deal of room left in identifying metaphorical instances and avoiding false positives. This task is confounded by the imbalance in the presence of metaphors and literal usage in our corpus. 

\begin{center}
    \begin{figure}[h!]
        \centering
        \includegraphics[width=0.5\linewidth]{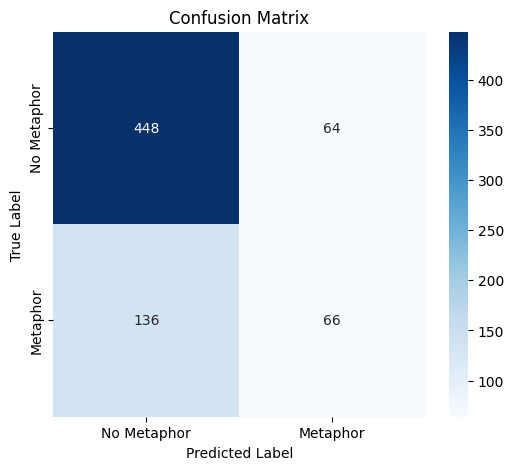}
        \caption{Confusion matrix of metaphor detection}
        \label{fig:cm}
    \end{figure}
\end{center}

A confusion matrix was generated as shown in figure \ref{fig:cm}, providing a visual breakdown of true positives, true negatives, false positives, and false negatives, which helps to understand the types of errors made by the model.
Given the potential imbalance between 'Metaphor' and 'No Metaphor' classes (5115 vs 2022 sentences as per initial stats), techniques like oversampling the minority class (metaphor), undersampling the majority class, or using class weights during training can be explored.

\begin{center}
    \begin{figure}[h!]
        \centering
        \includegraphics[width=0.5\linewidth]{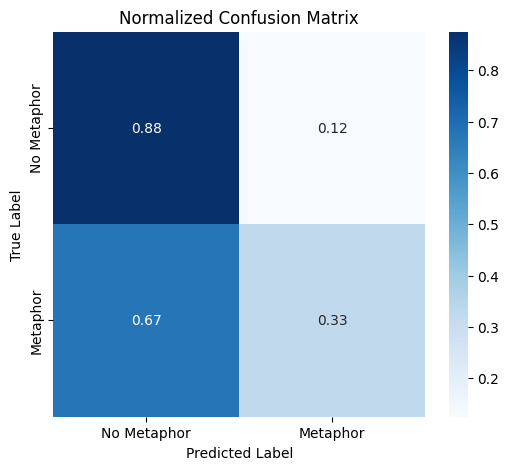}
        \caption{Normalized Confusion matrix of metaphor detection.}
        \label{fig:confusion matrix}
    \end{figure}
\end{center}

A normalized confusion matrix was generated as shown in figure \ref{fig:confusion matrix} suggests how well the model distinguishes between Metaphor and No Metaphor class. After normalization, each row represents proportions instead of raw counts, allowing for easier comparison across classes even when the HiLeMe corpus is imbalanced.

\subsection{Misclassified Sentences Analysis}
False Positives are instances where the model predicted a metaphor, but the sentence was actually non-metaphorical. Analyzing these can help understand what non-metaphorical phrases might be mistakenly interpreted as metaphors by the model.
False Negatives are instances where the model failed to detect an actual metaphor. Examining these cases can reveal patterns in metaphors that the model struggles to identify.

\begin{center}
    \begin{figure}[h!]
        \centering
        \includegraphics[width=0.8\linewidth]{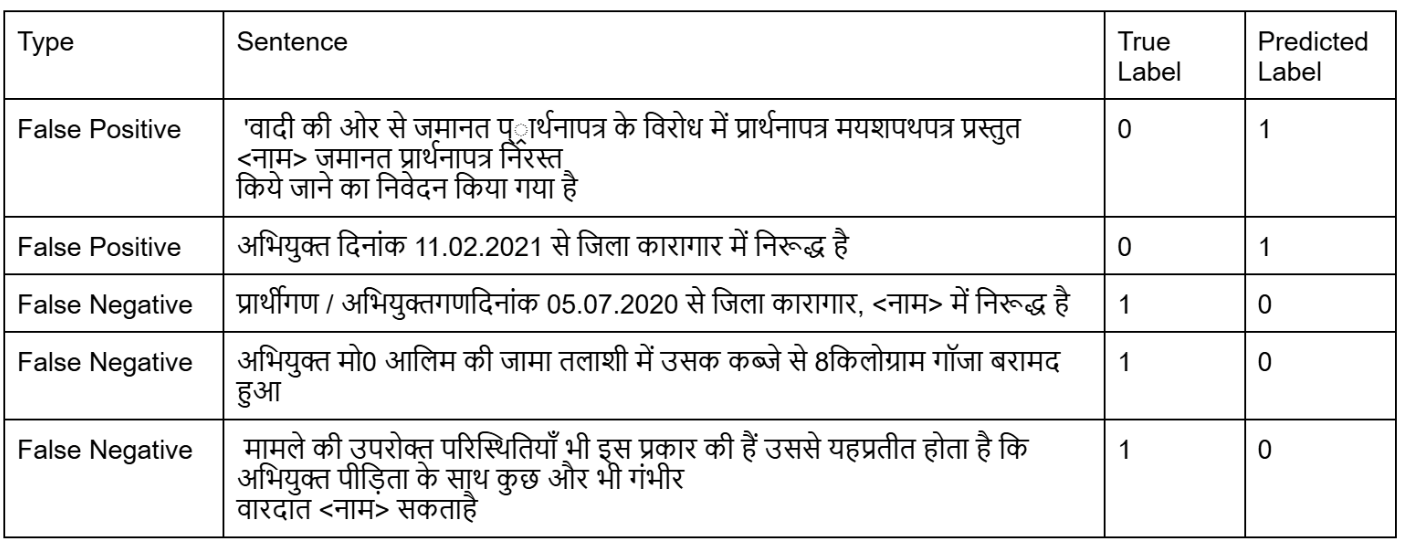}
        \caption{Examples of misclassified sentences }
        \label{fig:placeholder}
    \end{figure}
\end{center}

Examples of misclassified sentences as shown in figure \ref{fig:placeholder}, including both false positives (predicted metaphor, actual no metaphor) and false negatives (predicted no metaphor, actual metaphor), were successfully identified, offering specific instances of model error.

\subsection{Discussion and way forward}
Metaphor detection in legal discourse has invited treatments from qualitative and quantitative approaches. While qualitative approaches from linguistics have helped identify the various types of metaphors in use in languages \citep{lakoff1981}, computational and model implementations have been fraught with challenges. These challenges have stemmed from the intrinsic difficulty in identifying and detecting metaphorical and figurative usage in languages. These latter challenges have been addressed through generalized schemata such as the MIPVU \cite{steen2010}. On the other hand, model based approaches have relied on a plethora of methods, be they AN pair classification, both through supervised and unsupervised methods, cosine distances and other vector based methods. BERT based methods have most certainly shed light on the representational space of metaphors but this has been so far restricted to oft researched languages. 

Under-studied and lesser-known languages pose yet another functional challenge in that they often do not have the labeled textual resources on which models could be trained and also, an extrinsic issue, namely that metaphors, being as they are harder to interpret do not often lend themselves easily to analytical methods that could be language agnostic. Efforts such as by \cite{aghazadeh-etal-2022} certainly lead the way in language agnostic methods to identify metaphors, they are still limited by the availability of labeled language resources.

Legal language and discourse, as outlined in section \ref{sec:qa}, pose yet another set of challenges that are somewhat orthogonal to mere identification of metaphorical usage in legal discourse. However, building parallel corpora and training on them with transformer models have shown encouraging outcomes \citep{sanchez-boyona2025}. \cite{barman2023} following the release of HLDC \cite{kapoor-etal-2022-hldc} have shown that bail judgments can be successfully classified to an accuracy of 94\% using transformer based architecture. 

In this paper, we advance the idea that HiLeMe offers a first pass solution to solving the metaphor identification problem in Hindi legal data. Our results indicate that 71\% accuracy can be achieved with mBert transformers. However, the precision, recall and F1-scores do underscore the need to assess how these models could be made to cross the hurdle of linguistic complexity, the ubiquity of metaphorical usage and them being identified as such, especially in the specific case of legal discourse where the complexity is compounded by jurisprudential discretion to use metaphors in the way judges see fit. The way forward, in the present context, would entail injecting more contextual information, be they in terms of cosine distances of word vectors, greater interpretability of these models - for greater adoption by juridical systems and extending these models to be implemented over parallel corpora akin to \cite{sanchez-boyona2025} among the scheduled languages of India and with greater optimism in a largely language agnostic way.

\section*{Acknowledgments}
We are grateful to the DigiTS project at the University of Tartu for providing computational resources. We also express our gratitude towards the West Bengal National University of Juridical Sciences, Kolkata, India for their generous grant that made some of this research possible. This work was also made possible by Jadavpur University's favorable resources to foster interdisciplinary research.

\bibliography{metaphor}
\bibliographystyle{plainnat}
\end{document}